\documentclass[sigconf]{acmart}

\usepackage{colortbl}
\usepackage{type1cm}
\usepackage{flushend}
\usepackage{bm}
\usepackage{color}
\usepackage{balance}

\newcommand{\etal}{\textit{et al}.}
\newcommand{\eg}{\textit{e.g.}, }
\newcommand{\ie}{\textit{i.e.}, }
\newcommand{\argmax}{\mathop{\rm arg~max}\limits}

\AtBeginDocument{%
 \providecommand\BibTeX{{%
 \normalfont B\kern-0.5em{\scshape i\kern-0.25em b}\kern-0.8em\TeX}}}

\copyrightyear{2021}
\acmYear{2021}
\setcopyright{rightsretained}
\acmConference[MM '21] {Proceedings of the 29th ACM Int'l Conference on Multimedia}{October 20--24, 2021}{Virtual Event, China.}
\acmBooktitle{Proceedings of the 29th ACM Int'l Conference on Multimedia (MM '21), Oct. 20--24, 2021, Virtual Event, China}
\acmPrice{}
\acmISBN{978-1-4503-8651-7/21/10}
\acmDOI{10.1145/3474085.3475557}

\begin{document}
\fancyhead{}

\title[Sensor-Augmented Egocentric-Video Captioning]{Sensor-Augmented Egocentric-Video Captioning with Dynamic Modal Attention}

\author{Katsuyuki Nakamura}
\orcid{0000-0002-8074-2279}
\affiliation{%
 \institution{Hitachi, Ltd.}
 \city{Tokyo}
 \country{Japan}
 \postcode{185-8601}
}
\email{katsuyuki.nakamura.xv@hitachi.com}

\author{Hiroki Ohashi}
\orcid{0000-0001-6970-2412}
\affiliation{%
 \institution{Hitachi, Ltd.}
 \streetaddress{1-280, Higashi-koigakubo}
 \city{Tokyo}
 \country{Japan}
 \postcode{185-8601}
}
\email{hiroki.ohashi.uo@hitachi.com}

\author{Mitsuhiro Okada}
\affiliation{%
 \institution{Hitachi, Ltd.}
 \city{Tokyo}
 \country{Japan}
}
\email{mitsuhiro.okada.uf@hitachi.com}
\renewcommand{\shortauthors}{Nakamura, et al.}

\begin{abstract}
Automatically describing video, or video captioning, has been widely studied in the multimedia field. This paper proposes a new task of sensor-augmented egocentric-video captioning, a newly constructed dataset for it called MMAC Captions, and a method for the newly proposed task that effectively utilizes multi-modal data of video and motion sensors, or inertial measurement units (IMUs). While conventional video captioning tasks have difficulty in dealing with detailed descriptions of human activities due to the limited view of a fixed camera, egocentric vision has greater potential to be used for generating the finer-grained descriptions of human activities on the basis of a much closer view. In addition, we utilize wearable-sensor data as auxiliary information to mitigate the inherent problems in egocentric vision: motion blur, self-occlusion, and out-of-camera-range activities. We propose a method for effectively utilizing the sensor data in combination with the video data on the basis of an attention mechanism that dynamically determines the modality that requires more attention, taking the contextual information into account. We compared the proposed sensor-fusion method with strong baselines on the MMAC Captions dataset and found that using sensor data as supplementary information to the egocentric-video data was beneficial, and that our proposed method outperformed the strong baselines, demonstrating the effectiveness of the proposed method.
\end{abstract}




\begin{teaserfigure}
 \centering
 \includegraphics[width=0.95\textwidth]{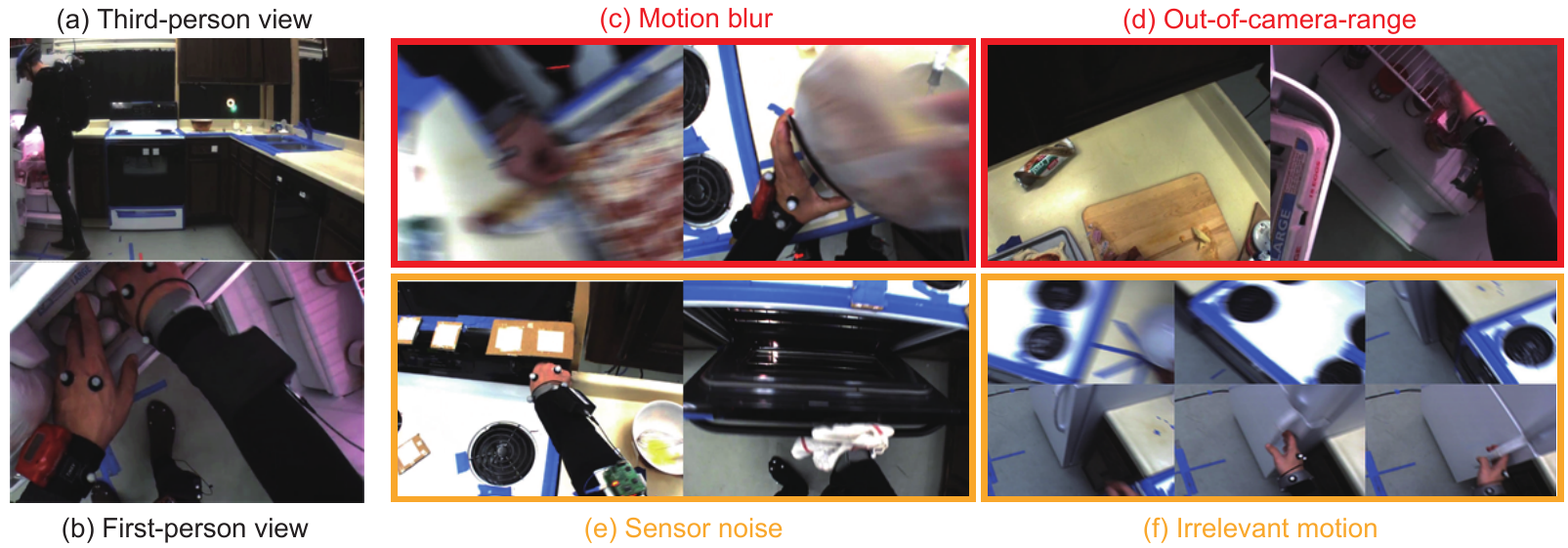}
 \caption{(a)--(b) First-person view is effective for observing more detailed hand operations compared with third-person view. (c)--(d) First-person view can be affected by harmful situations. (e) Sensor data contain noise caused by electromagnetic cooker. (f) The subject reaches a hand for the fridge while walking, where ``walking'' is the ground-truth activity. $\copyright$Ekaterina H. Spriggs.}
 \label{Fig_TeaserFigure}
\end{teaserfigure}

\maketitle
\section{Introduction}
\label{sec:introduction}
Video captioning is an active research topic in the multimedia field. Most studies on this topic have attempted to describe events using third-person video data~\cite{Chen2019}, but these studies have demonstrated the difficulty in describing finer-grained activities such as cooking because these activities require more detailed information than third-person video data. Egocentric video, which is recorded by using a wearable camera, is useful for this purpose because it can produce much finer-grained information on the activities of the camera wearer on the basis of the closer view (Figures \ref{Fig_TeaserFigure} (a)--(b)). However, video captioning based on the egocentric video is not trivial because it, in nature, often suffers from motion blur, self-occlusion, and out-of-camera-range activities (Figures \ref{Fig_TeaserFigure} (c)--(d)). It is therefore better to utilize supplementary information to overcome these difficulties.

This paper proposes a new task of sensor-augmented egocentric-video captioning for the finer-grained description of human activities. It utilizes egocentric-video data along with motion data from wearable sensors, or inertial measurement unit (IMU) sensors. For this task, we constructed a new dataset called MMAC Captions\footnote{The dataset will be released to facilitate further research.} by extending the CMU-Multimodal Activity (CMU-MMAC) dataset~\cite{Spriggs2009}. MMAC Captions contains 16 hours of egocentric-video and wearable-sensor data with more than 5,000 egocentric activity descriptions.

One of the difficulties of the proposed task is how to fuse the differing video and sensor modalities. Although both of these complementary modalities are useful in many situations, one modality could have a negative impact on performance. For example, sensor data may occasionally contain undesirable noise caused by surrounding objects or by irrelevant motions to the target activity (Figures \ref{Fig_TeaserFigure} (e)--(f)). In such cases, not using sensor data or assigning less weights to it is preferable. To this end, we propose a dynamic modal attention (DMA) module, which determines the modalities to be emphasized more (or less) depending on the context.

We compared the proposed sensor-fusion method with strong baselines on the MMAC Captions dataset. The experimental results showed that the sensor data was beneficial as supplementary information to the egocentric video data, and that our DMA module effectively determined the modalities to be emphasized.

The key contributions of this study are summarized as follows. (1) We propose a new task of sensor-augmented egocentric-video captioning for the finer-grained description of human activities, and provide a newly constructed MMAC Captions dataset that contains 16 hours of egocentric video and sensor data with more than 5,000 egocentric activity descriptions. (2) We propose a DMA module that determines the modalities to be emphasized more (or less) depending on the contexts. (3) We experimentally demonstrate the effectiveness of using sensor data and that of the DMA module through comparisons with strong baseline methods.

This paper is structured as follows. Section \ref{sec_related_work} reviews existing research on activity recognition. Section \ref{sec_dataset} introduces our MMAC Captions dataset. Sections \ref{sec_method} and \ref{sec_experiment} describe the method for sensor-augmented video captioning and its performance evaluation, respectively.

\section{Related Work}
\label{sec_related_work}
\paragraph{\bf{Image and Video Captioning}} There has been much research on generating textual descriptions from images and videos~\cite{Chen2019}. One approach is phrase-based captioning, in which an image is first converted into several phrases using object recognition and action recognition, then connecting them with textual modeling~\cite{Ushiku2011,Ordonez2011,Kulkarni2013,Toderici2010,Guadarrama2013,Rohrbach2013,Kuznetsova2012,Ushiku2015,Lebret2015}. This approach can avoid critical mistakes because it uses a prefixed word sequence; however, it is difficult to generate natural sentences. Another approach is to use an encoder-decoder model for image and video captioning~\cite{Karpathy2014,Vinyals2015,Chen2015,Mao2015,Liu2016,Xu2015,Wu2016,Venugopalan2015,Donahue,Venugopalan2016,Yao2015,Baraldi2017,Pan2016,Hendricks2016,Krishna2017}. The pioneering work by Venugopalan~\etal~\cite{Venugopalan2015} uses a sequence-to-sequence model that combines a convolutional neural network (CNN)-based image encoder and long short-term memory (LSTM)-based textual decoder. Many studies have proposed extensions to this work from various perspectives to improve performance; selective encoding~\cite{Chen2018}, boundary-aware encoding~\cite{Baraldi2017}, adaptive attention~\cite{Lu2017}, word-transition molding~\cite{Chen2018b,Ke2019}, and object relational modeling~\cite{Zhang2020,Aafaq2019,Zhou2019,Pan2020}. Transformer~\cite{Ashish2017} has been proven to be more effective for video captioning~\cite{Chen2018c,Zhou2018,Sun2019,Pan2020,Lei2020,Ging2020,Jin2020}. Zhou~\etal~\cite{Zhou2018} first proposed the end-to-end transformer model for dense video captioning. Lei~\etal~\cite{Lei2020} introduced a memory-augmented recurrent transformer based on bidirectional encoder representations from transformers (BERT)~\cite{Devlin2019}.

\paragraph{\bf{Multi-modal Captioning}} Several works have explored captioning based on multi-modal information such as that from video and audio information~\cite{Ramanishka2016}, and multivariate well logs~\cite{Tong2017}. One of the typical methods for combining multi-modal information is to use multi-modal attention~\cite{Rahman2019,Hori2017,Xu2017}. The attention mechanism has been further explored with the transformer architecture for multi-modal video captioning. Iashin~\etal ~proposed a feature transformer for video and audio input~\cite{Iashin2020} and bi-modal attention for both modalities~\cite{Iashin2020a}.

Our work expands upon these prior works and exploits the attention-based modality-integration module for dynamically changing emphasis on different video and sensor modalities depending on the contexts.

\paragraph{\bf{Activity Recognition using Wearable Camera and Sensors}} There is a substantial body of work on egocentric vision, ranging from activity recognition to video summarization~\cite{Nguyen2016,Betancourt2015}. One of the following cues is commonly used in activity recognition; a motion cue, which uses coherent motion patterns for egocentric video~\cite{Kitani2011,Poleg2015,Ryoo2015,Yonetani2016,Singh2017}, an object cue, which uses objects in video sequences~\cite{Wang2020a, Hamed2012,Fathi2013,Fathi2011a,Lee2012,Fathi2011,Damen2014}, or an integrated cue, which uses both motion and object information~\cite{Li2013,Fathi2012,Ohnishi2015a,Ma2016,Li2015,al2018hierarchical}. Much effort has been put into constructing applications~\cite{Nagarajan2020,Ng2020,Yuan2019,Fan2016,Bolanos2018,Ohnishi2015a} and datasets~\cite{Damen2018,Spriggs2009,Nakamura2017,Kong2019,Shan2020}, too.

Several studies have used non-visual wearable sensors for activity recognition, including accelerometers and heart rate sensors~\cite{Nakanishi2015,Nakamura2017,ohashi2017augmenting,ohashi2018attributes}. Bao~\etal~\cite{Bao2004} used multiple accelerometers on the hip, wrist, arm, ankle, and thigh to classify 20 classes of household activities. Spriggs~\etal~\cite{Spriggs2009} used an egocentric camera, IMUs, and other sensors to classify 29 classes of kitchen activities. Maekawa~\etal~\cite{Maekawa2010} used a wrist-mounted camera and sensors to detect activities in daily life. The present study extends these prior works to tackle a newly proposed task of sensor-augmented egocentric-video captioning.

\section{MMAC Captions Dataset}
\label{sec_dataset}
\begin{figure*}[!t]
\centering
\includegraphics[width=0.95\textwidth]{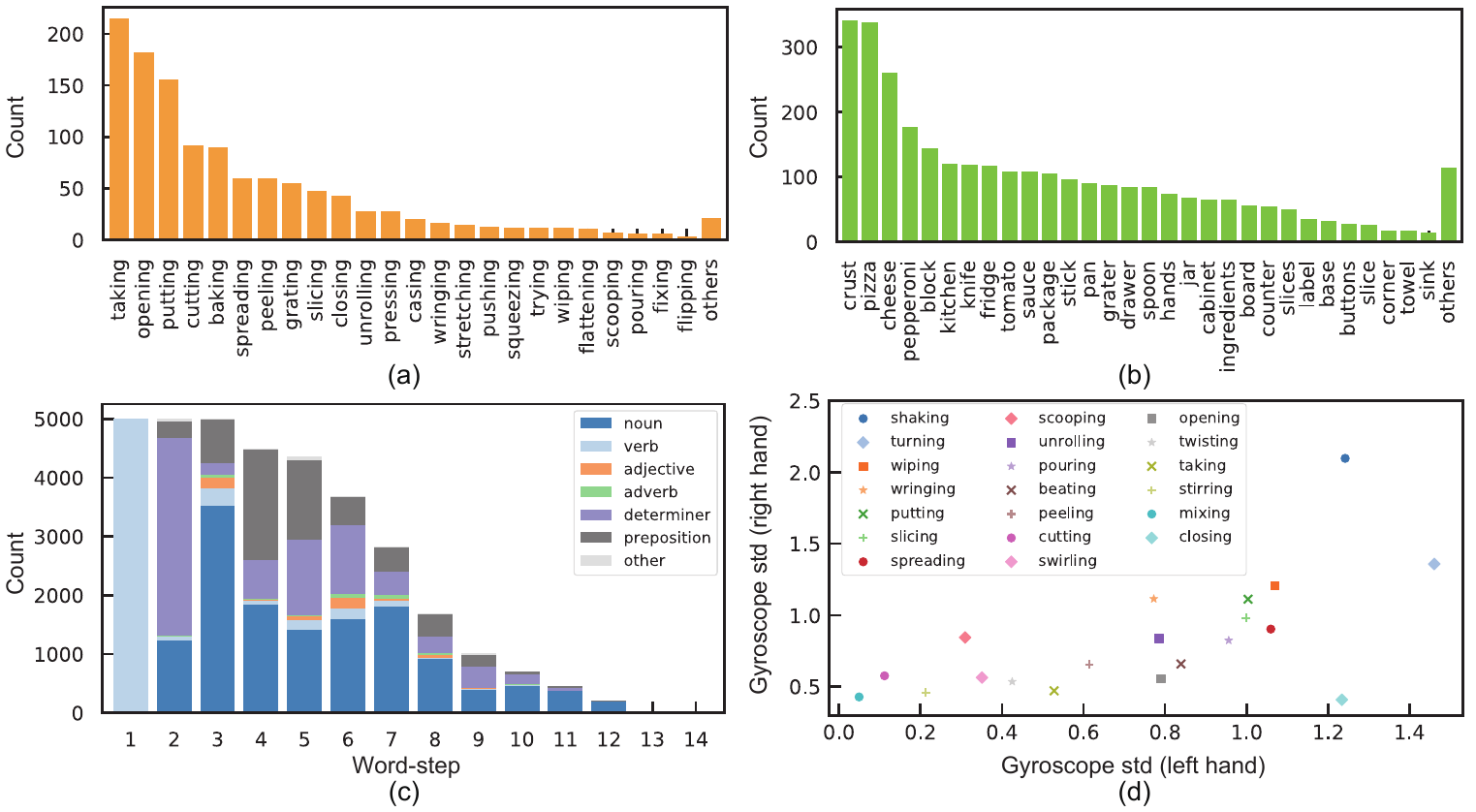}
\caption{Statistics of MMAC Captions: (a) distribution of verb classes, (b) distribution of noun classes, (c) distribution of word-types at each word step, and (d) the scatter plot for standard deviations of gyroscope in both hands.}
\label{Fig_Dataset_stats}
\end{figure*}

The CMU-MMAC dataset~\cite{Spriggs2009} is the first dataset that augments egocentric video with wearable IMU data. The sampling rate is 30 fps with a resolution of both 800$\times$600 and 1024$\times$768 for egocentric video and a maximum of 125 Hz for 9-axes IMUs (a 3-axes gyroscope, 3-axes accelerometer, and 3-axes compass). 9 IMUs were attached to the subject’s body: both forearms/upper arms, both thighs/lower limbs, and back. A Point Grey FL2-08S2C-C camera was used for capturing egocentric video. To construct this dataset, 43 participants wore a wearable camera and IMU sensors and performed five types of cooking activities in a kitchen (\ie {\it Brownie}: making brownies, {\it Salad}: preparing a salad, {\it Pizza}: making pizza, {\it Eggs}: frying eggs, and {\it Sandwich}: making a sandwich). Although several types of annotations including object and action classes are provided\footnote{\url{http://kitchen.cs.cmu.edu/.}}, detailed activity descriptions are not.

We therefore introduce a new dataset called \textbf{MMAC Captions}, which contains 5,002 activity descriptions for 16 hours of egocentric data. Two annotators were independently asked to first define a ``{\it segment}'', which corresponds to a short period that contains only one activity, by determining the timestamps of the start and end of the segment. Second, they were asked to provide one sentence per segment. Finally, cross checking was carried out to integrate multiple captions into one consistent sentence after each annotator finished their own annotations. We omit the subject of the sentences in the annotations because the subject is always the person who wears the sensors and camera. The dataset will be made publicly available.

A number of activity description examples are shown as follows. \\
- {\it Spreading tomato sauce on pizza crust with a spoon.} \\
- {\it Taking a fork, knife, and peeler from a drawer.}\\
- {\it Cutting a zucchini in half with a kitchen knife.} \\
- {\it Moving a paper plate slightly to the left.} \\
- {\it Stirring brownie batter with a fork.} \\
We found that the provided annotations were much more diverse than those of the action or object categories in a sense that there exist multiple objectives, adverbs, and a wide distribution of sentence lengths.

We used the Natural Language Toolkit (NLTK)\footnote{\url{https://www.nltk.org/}} and TreeTagger\footnote{\url{https://www.cis.uni-muenchen.de/~schmid/tools/TreeTagger/}} to analyze the dataset, resulting in observing 131 noun classes and 89 verb classes as shown in Figures \ref{Fig_Dataset_stats} (a) and (b). Figure \ref{Fig_Dataset_stats} (c) shows the distribution of word-types at each word step. All the sentences start with a verb, and a determiner and noun pair follows in many cases. The average sentence length is seven words, and the maximum length is 14 words. The average duration of a segment is 6.7 seconds, and the median is 3.0 seconds. Figure \ref{Fig_Dataset_stats} (d) shows that sensor data from an IMU, such as the gyroscope, are useful to predict activities. For example, {\it shaking} and {\it turning} take a significantly higher variance than other activities, suggesting that IMU signals are effective for activity description.

\begin{figure*}[t]
\centering
\includegraphics[width=0.90\textwidth]{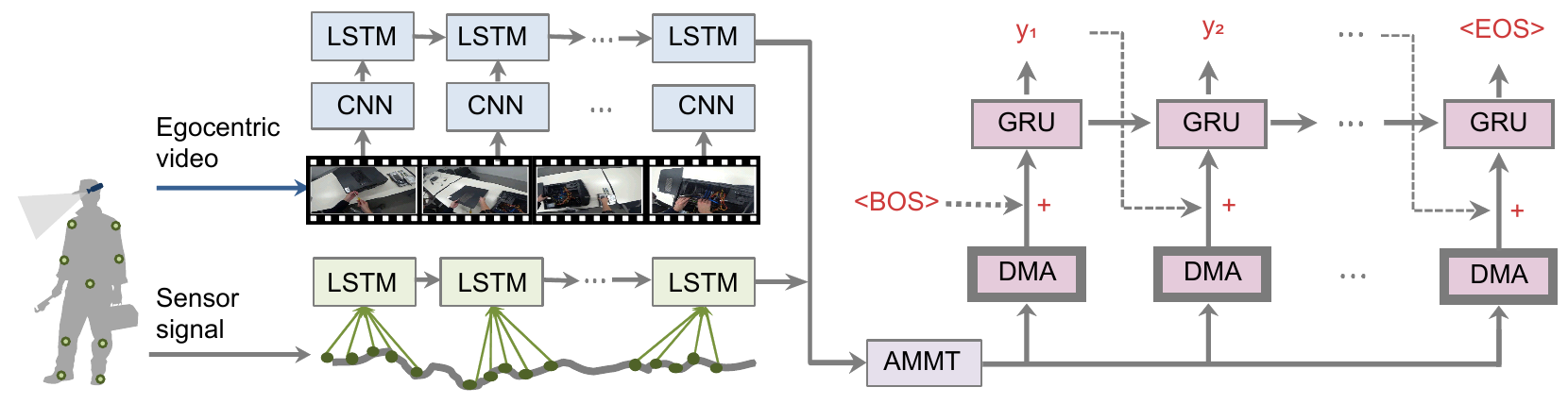}
\caption{Flow of proposed approach, which involves inputting egocentric video and sensor signals, and outputting the activity description from an egocentric perspective. Visual and sensor encoders individually output their representations. The AMMT module adaptively integrates these representations. A DMA module dynamically determines the best fusion of different representations for generating each word in a sentence.}
\label{Fig_Overview}
\end{figure*}

\section{Method}
\label{sec_method}
The flow of our approach is shown in Figure~\ref{Fig_Overview}. Our goal is to describe activities using egocentric-video and wearable-sensor data. To achieve this goal, we extend the boundary-aware neural encoding model~\cite{Baraldi2017} for multi-modal activity descriptions.

The core idea includes two key aspects. First, we model a sensor encoder with recurrent neural networks and introduce a learnable transformation module to effectively integrate multi-modal features. Second, we propose a step-wise modality-attention module to dynamically determine the best fusion of different representations for generating each word in a sentence.

\subsection{Multi-modal Encoder}
\label{sec_multimodal_encoder}
The input to our model is a sequence of video frames $\mathcal{V}=\{\mathbf{v}_1,\mathbf{v}_2, \\ ..., \mathbf{v}_K\}$ and wearable-sensor signals $\mathcal{S}=\{\mathbf{s}_1,\mathbf{s}_2,...,\mathbf{s}_T\}$, where $\mathbf{v}_{t}$ is a video frame at time $t$, and $\mathbf{s}_{t}$ is a 63-dimensional vector at time $t$ from 9-axis IMUs attached at seven locations on the body\footnote{There were originally nine locations, but the data collected from two sensors were incomplete, and were therefore not included in the CMU-MMAC dataset.}.

\paragraph{\bf{Visual Encoder}} We developed a visual encoder $\phi_{\rm V}$ based on the boundary-aware neural encoder~\cite{Baraldi2017} that can identify discontinuity points between frames such as appearance change and motion change. Our intuition is that such discontinuity frequently occurs in egocentric videos because there are many undesirable imaging situations such as motion blur, frame out, illuminance change, and self-occlusion. Therefore, we assumed a boundary detector would work efficiently. Given a video sequence, we compute a visual representation as $\mathbf{h}_{\rm V}=\phi_{\rm V}(\mathcal{V})$, where $\mathbf{h}_{\rm V}$ represents the final hidden state of the LSTM layer.

\paragraph{\bf{Sensor Encoder}} Unlike in $\phi_{\rm V}$, we do not use the boundary detector in sensor encoder $\phi_{\rm S}$. In many cases, boundary detection occurs too often during the sensor-encoding step, which results in an enormous state initialization. We therefore developed a neural encoder based on the LSTM layers without boundary-aware encoding. Raw sensor signals are first resampled at 30 Hz using a piecewise linear interpolation and directly fed into the LSTM input layer. With the sensor signals, we compute sensor representation as $\mathbf{h}_{\rm S} =\phi_{\rm S} (\mathcal{S})$, where $\mathbf{h}_{\rm S} $ is also the final hidden state of the LSTM layer.

\paragraph{\bf{Asymmetric Multi-modal Transformation (AMMT)}} Once the representations of video and sensor have been encoded, the decoder can generate a textual description by simply concatenating both representations (\ie $\mathbf{h}_{\rm V+S} = \mathbf{h}_{\rm V} \oplus \mathbf{h}_{\rm S})$.

However, the simple concatenation may not be optimal because the amount of useful information contained in each representation vector may be significantly different due to the difference in their dimensions, or the length of the representation vectors. To mitigate the possible imbalance, we introduce a learnable transformation module called the asymmetric multi-modal transformation (AMMT), which is inspired by feature-wise linear modulation~\cite{Perez2018}. AMMT is defined as follows:
\begin{equation} \label{eq_affine}
\mathbf{h}_{\rm V+S} = \mathbf{h}_{\rm V} \oplus \left( \mathbf{W}_{c} \mathbf{h}_{\rm S} + \mathbf{b}_{c} \right),
\end{equation}
where $\mathbf{W}_{c}$ is the weight matrix representing a linear transformation and $\mathbf{b}_{c}$ is a bias. The $\mathbf{W}_{c}$ and $\mathbf{b}_{c}$ are learnable vectors, which are initially set to identity transformation (\ie $\mathbf{W}_{c}=\mathbf{I}$ and $\mathbf{b}_{c}=0$). In other words, we use a standard concatenation in the initial phase and update parameters during the training phase.

The intuition behind the design of asymmetry, or applying linear transformation only to $\mathbf{h}_{\rm S}$, is two-fold. First, the mitigation of the aforementioned possible imbalance can be achieved by applying the transformation to one of the representations rather than both; the latter has a risk of over-fitting due to the redundant parameters. Second, because the sensor data sometimes contain undesirable noise, which is explained in Section~\ref{sec:introduction}, adjusting the sensor representation so that it does not adversely affect the performance is preferable. In addition to this integrated representation $\mathbf{h}_{\rm V+S}$, AMMT forwards $\mathbf{h}_{\rm V}$ and $\mathbf{h}_{\rm S}$ as well because there are a number of cases where using only single modality is preferable (\eg sensor data containing undesirable noise).

\subsection{Multi-modal Decoder}
\label{sec_decoder}
\paragraph{\bf{Sentence Generation}} Given the aforementioned representations from the multi-modal encoder, our decoder generates a sentence $\mathcal{\hat{Y}}=\{\hat{\mathbf{y}}_{0},\hat{\mathbf{y}}_{1},\hat{\mathbf{y}}_{2},...,\hat{\mathbf{y}}_{N}\}$, where $\hat{\mathbf{y}}_{i}$ is a word with one-hot-vector encoding at step $i$. The objective function for sentence generation is
\begin{equation}
\theta^{*} = \argmax_{\mathbf{\theta}} \sum_{i=1}^{N} \log \Pr \left( \hat{\mathbf{y}}_{i} | \hat{\mathbf{y}}_{i-1}, \hat{\mathbf{y}}_{i-2}, ...,\hat{\mathbf{y}}_{0}, \mathbf{z}; \theta \right),
\end{equation}
where $\theta$ denotes all parameters in the model, and $\mathbf{z}$ is a set of representations from the encoder. We compute the probability of a word as follows:
\begin{equation}
\Pr \left( \hat{\mathbf{y}}_{i} | \hat{\mathbf{y}}_{i-1}, \hat{\mathbf{y}}_{i-2}, ...,\hat{\mathbf{y}}_{0}, \mathbf{z}; \theta \right) \propto \hat{\mathbf{y}}_{i}^{\mathsf{T}} \text{softmax}(\mathbf{W}_{h} \mathbf{h}_{i}),
\end{equation}
where $\mathbf{W}_{h}$ is a matrix to convert a hidden state to the same dimension as word vector $\hat{\mathbf{y}}_{i}$, and $\mathbf{h}_{i}$ is an output hidden state of decoder at word step $i$ by using gated recurrent unit (GRU). We use the special token \textless BOS\textgreater\space as the initial word $\hat{\mathbf{y}}_{0}$ and \textless EOS\textgreater\space as the termination symbol, which enables variable-length sentences to be generated.

\paragraph{\bf{Dynamic Modal Attention (DMA)}} \label{sec_DMA} DMA takes representation vectors given by AMMT as input, and outputs the best-integrated features at each decoding step on the basis of attention on different types of representations.

Given representations $\mathbf{z} = (\mathbf{h}_{\rm V},\mathbf{h}_{\rm S},\mathbf{h}_{\rm V+S})^{\mathsf{T}}$ from AMMT, DMA takes the form of the weighted sum of each representation as follows.
\begin{equation} \label{eq_gumbel_approx}
g(\mathbf{z}) = \sum_{k}{\zeta_{k, i} \mathbf{W}_{k} \mathbf{h}_{k}},
\end{equation}
where $k \in \{{\rm V}, {\rm S}, {\rm V+S}\}$, $\mathbf{W}_{k}$ are the linear transformations for adjusting the dimensions so that all representation vectors have the same length, and $\left\{\zeta_{k, i} \, | \, \zeta_{k} \in [0, 1], \sum_{k}\zeta_{k}=1\right\}$ are the weights of each representation at word step $i$.

There are multiple choices for the design of $\zeta_{k, i}$ including simple softmax, Gumbel softmax~\cite{Jang2017}, straight through (ST) Gumbel softmax~\cite{Jang2017}, those with temperature parameters, and weighted variants of those designs to incorporate some preference on a certain type of representation. Here we define the general form of these variants as follows, and give the ablation study in Section \ref{sec:analysis}.
\begin{equation} \label{eq_gumbel_zeta}
\zeta_{k, i} = \frac{\exp ((g_{k, i}+\log \pi_{k, i}) / \tau)}{\sum_{k'} \exp ((g_{k', i}+\log \pi_{k', i}) / \tau)},
\end{equation}
where $g_{k, i}=0$ in softmax, and $g_{k, i}=-\log(-\log u), u \sim U(0,1)$ in (ST) Gumbel softmax. $U(0,1)$ is the uniform distribution. When we use ST Gumbel softmax, we can transform $\zeta_{k, i}$ into a one-hot vector in the forward pass by $\zeta_{k, i}'=\text{one-hot} \left( \argmax_{k'} \zeta_{k', i} \right)$, and approximate $\zeta_{k, i}'$ by $\zeta_{k, i}$ in the backward pass to enable back-propagation. $\tau$ is the temperature to control the ``hardness'' of the attention. A smaller $\tau$ tends to give ``harder'' attention, which results in making an output vector close to a one-hot one. $\pi_{k}$ are defined as follows.
\begin{equation}
\pi_{k, i}
=\frac{c_{k}\rho_{k, i}}{\sum_{k'}c_{k'}\rho_{k', i}},
\end{equation}
where $c_{k}$ are the weights that represent the preference for each representation. $\rho_{k, i}$ are defined as follows.
\begin{equation} \label{rho_v}
\rho_{k, i} = \frac{\eta_{k, i}}{\sum_{k'}\eta_{k', i}},
\end{equation}
where $\eta_{k,i} = \text{sigmoid} \left( \mathbf{W'}_{k, i} \mathbf{h}_{k} + \mathbf{b}_{k,i} \right)$. Note that we calculate the attention $\zeta_{k, i}$ at each word step $i$, which means that DMA can dynamically and flexibly change the attention at each word step in a sentence depending on the context. See Section \ref{sec:analysis} for the ablation study on the different choices of $g_{k, i}$, $\tau$, and $c_{k}$.

\begin{table*}[!t]
\begin{center}
\vspace{-2mm}
\caption{Captioning performance on MMAC Captions dataset. V indicates video data and S are sensor data. B-N represents BLEU score at N-gram. We computed the averaged value of each metric with six training runs.}
\label{Tab_results_seen}
\begin{tabular}{@{}lm{1.3cm}m{1.3cm}m{1.3cm}m{1.3cm}m{1.3cm}m{1.3cm}m{1.3cm}m{1.3cm}@{}}
\toprule
Method & B-1 & B-2 & B-3 & B-4 & B-5 & METEOR & CIDEr-D & SPICE \\ \midrule
LSTM-YT~\cite{Venugopalan2015a} & 52.9 & 45.8 & 39.8 & 34.6 & 30.7 & 24.8 & 292.9 & 0.423 \\
S2VT~\cite{Venugopalan2015} & 64.5 & 58.0 & 52.7 & 48.5 & 45.6 & 33.8 & 428.9 & 0.557 \\
ABiViRNet~\cite{Bolanos2018} & 62.4 & 55.6 & 50.3 & 45.9 & 42.8 & 32.4 & 419.4 & 0.557 \\
BA~\cite{Baraldi2017} & 66.8 & 60.5 & 55.6 & 51.6 & 48.8 & 35.8 & 460.3 & 0.597 \\
VTransformer~\cite{Zhou2018} & 67.9 & 61.8 & 56.7 & 52.5 & 49.6 & 36.5 & 471.4 & \textbf{0.625} \\ \midrule
Ours (V) & 66.8 & 60.5 & 55.6 & 51.6 & 48.8 & 35.8 & 460.3 & 0.597 \\
Ours (S) & 49.5 & 41.2 & 35.3 & 30.9 & 27.8 & 23.0 & 278.7 & 0.384 \\
Ours (V+S fixed) & 67.6 & 61.5 & 56.6 & 52.6 & 49.8 & 36.5 & 476.9 & 0.613 \\
Ours (V+S dynamic) & \textbf{68.4} & \textbf{62.3} & \textbf{57.4} & \textbf{53.5} & \textbf{50.7} & \textbf{37.2} & \textbf{484.8} & 0.618 \\ \bottomrule
\end{tabular}
\end{center}
\end{table*}

\vspace{-2mm}
\section{Experiment}
\label{sec_experiment}
\subsection{Setup}
We evaluated our approach on the MMAC Captions dataset. We split the dataset into training, validation, and test sets so that each set includes all recipe types, resulting in 2,923, 838, and 1,241 data for the training, validation, and test sets, respectively.

\paragraph{\bf{Implementation Details}} We set the maximum step for video frames ($K$ for $\phi_{\rm V}$) and IMU signals ($T$ for $\phi_{\rm S}$) to be 80 and 240, respectively. If the number of video frames did not match $K$, we down-sampled or duplicated the video frame. We also down-sampled IMU signals to match $T$.

We used the VGG16 model pre-trained with ImageNet for $\phi_{\rm V}$. The representations in the encoder had the following dimensions: $\mathbf{h}_{\rm V} \in \mathbb{R}^{500}$, $\mathbf{h}_{\rm S} \in \mathbb{R}^{120}$, and $\mathbf{h}_{\rm V+S} \in \mathbb{R}^{620}$. Training was conducted by minimizing the cross-entropy loss with the Adam optimizer. We set the batch size to 100 sequences and the learning rates to $3 \times 10^{-4}$ (0 $\leq$ epoch $<$300), $1 \times 10^{-4}$ (epoch $<$400), and $5 \times 10^{-5}$ (400 $\leq$ epoch). The maximum number of words in the generated sentence, \ie $N$ in the decoder, were set to 15. We used Gumbel softmax for Eq. (\ref{eq_gumbel_zeta}), and set $\tau=0.05$ and $\left( c_{\rm V}, c_{\rm S}, c_{\rm V+S} \right) = \left( 1, 1, 1.5 \right)$ unless otherwise stated. During training, we used scheduled sampling~\cite{Bengio2015}, which improves generalization performance. PyTorch was used for the implementation.

\paragraph{\bf{Metrics and Baselines}} We used four metrics for the evaluation: BLEU, METEOR, CIDEr-D, and SPICE. BLEU is a precision metric of word n-grams between prediction and ground-truth sentence. METEOR is a more semantic metric that absorbs subtle differences in expression using a WordNet synonym. CIDEr-D measures the cosine similarity between a generated sentence and ground-truth sentence using term-frequency inverse-document-frequency and avoids the gaming effect with a stemming technique. SPICE is a metric that has shown correlations with human judgement. As in most previous studies, we used METEOR and CIDEr-D as the main metrics. They are computed using the MS-COCO caption evaluation tool~\cite{Chen2015a}. We used average metrics of 6-times training to avoid the effect of random initialization.

We chose reasonable, versatile baselines to verify the performance of the novel multi-modal captioning task:
\begin{enumerate}
\item{Egocentric-based method}: ABiViRNet~\cite{Bolanos2018}
\item{RNN-based method}: LSTM-YT~\cite{Venugopalan2015a}, S2VT~\cite{Venugopalan2015}, and BA~\cite{Baraldi2017}.
\item{Transformer-based method}: VTransformer~\cite{Zhou2018}.
\end{enumerate}

\subsection{Results}
\paragraph{\bf{Quantitative Results}} Table \ref{Tab_results_seen} lists the results on the MMAC Captions dataset. The proposed approach, which dynamically integrates vision and sensor data (Ours (V+S dynamic)) outperformed all the baselines. It achieved +1.4\ gain in METEOR compared with the vision-only result (Ours (V)). As METEOR is close to a subjective evaluation, the results justify our claim that sensor-augmented multi-modal captioning is effective for generating detailed activity descriptions. Our approach when using only the sensors (Ours (S)) was inferior to the other baselines, but interestingly, it had comparable performance to LSTM-YT. This result also supports the effectiveness of sensor information. Ours (V+S fixed) is the result obtained by using $\mathbf{h}_{\rm V+S}$ all the time without using AMMT and DMA. Although it outperformed both of the single-modality variants (Ours (V) and Ours (S)) as well as achieved comparable performance to the best baseline model (VTransformer), our proposed approach with AMMT and DMA (Ours (V+S dynamic)) acquired an additional gain of 0.7 in METEOR. This result indicates that it is important to dynamically change the attention to the different modalities, and DMA effectively worked from this perspective.

\begin{figure*}[t]
\centering
\includegraphics[width=0.92\textwidth]{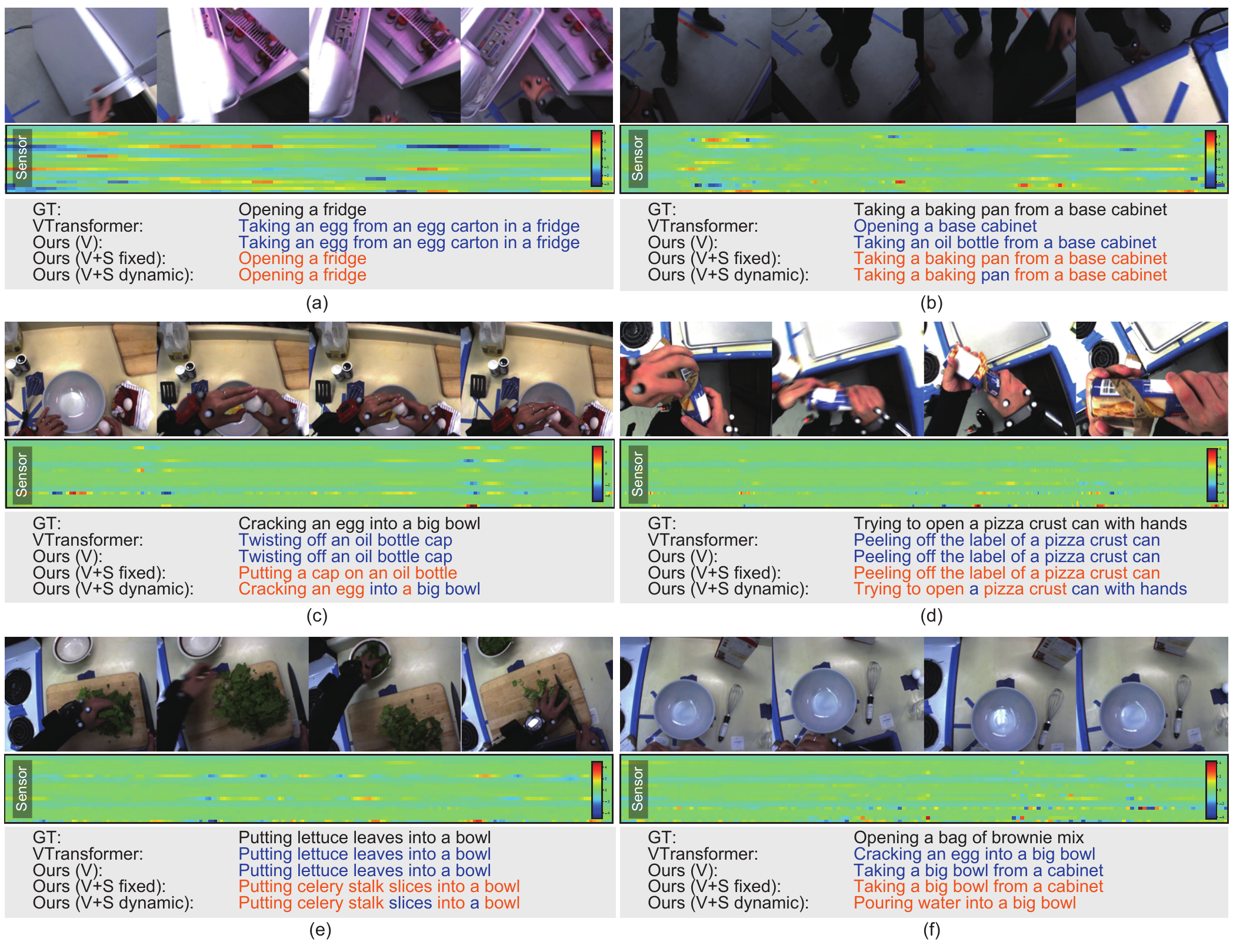}
\caption{Qualitative results. Notations: GT-Ground Truth, V-Vision, S-Sensor. Orange words represent those generated by attending multi-modal representation $\mathbf{h}_{\rm V+S}$, while blue ones represent those generated by $\mathbf{h}_{\rm V}$. $\copyright$Ekaterina H. Spriggs.}
\label{Fig_results}
\end{figure*}

\paragraph{\bf{Qualitative Results}} Qualitative results are shown in Figure \ref{Fig_results}. In each example, the top row is a snapshot of video frames. The middle row represents 21-dimensional gyro signals (3-axes by seven locations) arranged in chronological order. The bottom row shows the captioning results. We observed that the first words, (\ie verbs), are always generated by attending to a multi-modal representation $\mathbf{h}_{\rm V+S}$, confirming that sensor-augmented representation is especially effective for generating verbs.

Figures \ref{Fig_results} (a)--(b) indicate that sensor fusion helps infer the appropriate verbs and sometimes even nouns. While it is difficult for vision-only methods to distinguish visually similar actions such as {\it Opening} and {\it Taking} in these examples, the auxiliary sensor information was able to help generate the precise descriptions. Figures \ref{Fig_results} (c)--(d) show the effectiveness of our DMA module. The colors of the generated words indicate that DMA appropriately shifts the attention from V+S to V and vice versa when necessary, resulting in better captioning results. As seen in these examples, it sometimes attends to V rather than V+S when generating nouns. We believe this is reasonable because visual information should be especially important for nouns and noisy sensor information sometimes harms the performance, although it is generally helpful for generating verbs. We will further study these points in the final paragraph of this section, referring to Figure \ref{Fig_DMA_analysis}. Figure \ref{Fig_results} (e) represents a failure case where visual information alone is sufficient probably because the object is clear. In this case, sensor information caused some inaccurate inference (although it seems somewhat reasonable), but DMA failed to attend to V. Figure \ref{Fig_results} (f) is a challenging case, where an activity took place in the out-of-camera-range. In this case, all the methods including the proposed method failed to accurately generate the description. Generating an accurate description even in such a case would require the incorporation of much broader contexts in the long term, which will be our future work, potentially by exploiting transformer-like architecture.

\subsection{Further Analysis}
\label{sec:analysis}
\paragraph{\bf{Ablation Study}} We conducted an ablation study to investigate the characteristics of the proposed sensor-fusion method. In particular, we tested the following combinations: (i) {\it vision}: using only visual information, (ii) {\it vision+sensor}: multi-modal fusion using simple concatenation, (iii) {\it asymmetric fusion}: using AMMT for multi-modal fusion, (iv) {\it dynamic attention}: using DMA for sentence generation, and (v) {\it full}: using all the techniques.

As shown in Table \ref{Tab_Ablation}, {\it vision+sensor} outperformed {\it vision}, which suggests the effectiveness of the sensor modality as auxiliary information. We found fusion with only DMA without AMMT (iv) were worse than fusion with simple concatenation (ii). We assume this is because it is important to adjust the different representations, \ie $\mathbf{h}_{\rm V}$, $\mathbf{h}_{\rm S}$, and $\mathbf{h}_{\rm V+S}$, so that a possible imbalance among these representations can be calibrated before being input to DMA. Similarly, fusion with only AMMT (iii) resulted in a worse performance than fusion with simple concatenation (ii) probably because the transformation layer is unnecessary if DMA is not used, and the unnecessary redundant layer caused over-fitting. Finally, {\it full}, which uses both of AMMT and DMA, led to the best results.

\begin{table*}[t]
\begin{center}
\caption{Results of ablation study for multi-modal fusion.}
\label{Tab_Ablation}
\begin{tabular}{@{}m{0.9cm}llm{1cm}m{1cm}m{1cm}m{1cm}m{1cm}m{1cm}m{1.3cm}m{1.3cm}m{1.3cm}@{}}
\toprule
No & Fusion & AMMT & DMA & B-1 & B-2 & B-3 & B-4 & B-5 & METEOR & CIDEr-D & SPICE \\ \midrule
(i) & & & & 66.8 & 60.5 & 55.6 & 51.6 & 48.8 & 35.8 & 460.3 & 0.597 \\
(ii) & \checkmark & & & 67.6 & 61.5 & 56.6 & 52.6 & 49.8 & 36.5 & 476.9 & 0.613 \\
(iii) & \checkmark & \checkmark & & 67.5 & 61.3 & 56.4 & 52.5 & 49.7 & 36.4 & 474.0 & 0.607 \\
(iv) & \checkmark & & \checkmark & 67.0 & 60.8 & 55.9 & 52.0 & 49.2 & 36.1 & 467.7 & 0.602 \\
(v) & \checkmark & \checkmark & \checkmark & \textbf{68.4} & \textbf{62.3} & \textbf{57.4} & \textbf{53.5} & \textbf{50.7} & \textbf{37.2} & \textbf{484.8} & \textbf{0.618} \\ \bottomrule
\end{tabular}
\end{center}
\end{table*}

\paragraph{\bf{Analysis of AMMT}} We analyzed the design of AMMT in more detail. We conducted the ablation study by changing the symmetric and asymmetric patterns to see how the fusion module achieves the gain. The results presented in Table \ref{Tab_Affine} confirm the effectiveness of the asymmetric transformation. We assume this is because the possible imbalance due to the difference between visual and sensor representation is mitigated by applying a transformation. Introducing symmetric modulation, or applying transformations to both of the representations, decreased the performance probably because of the redundant parameters. Incorporating redundant parameters, despite being sufficient to apply a transformation to either of the representations, may lead to over-fitting. In contrast, both the asymmetric modulations improve the performance, especially when the transformation is applied to the sensor representations. One possible hypothesis is that the sensor data sometimes contain undesirable noise, which can be surpassed by applying a transformation.

\begin{table}[ht]
\begin{center}
\caption{Ablation study for AMMT. Linear ($\mathbf{h}_{\rm V}$/$\mathbf{h}_{\rm S}$) denotes the modality that linear transformation was applied in AMMT. The last row corresponds to Eq. (\ref{eq_affine}). B: BLEU, M: METEOR, C: CIDEr-D, S: SPICE. }
\label{Tab_Affine}
\begin{tabular}{@{}m{1.4cm}m{1.4cm}lllll@{}}
\toprule
Linear$(\mathbf{h}_{\rm V})$ & Linear$(\mathbf{h}_{\rm S})$ & B-1 & B-4 & M & C & S \\ \midrule
 & & 67.0 & 52.0 & 36.1 & 467.7 & 0.602 \\
\checkmark & \checkmark & 67.0 & 51.7 & 35.9 & 465.0 & 0.602 \\
\checkmark & & 67.4 & 52.4 & 36.2 & 468.4 & 0.604 \\
 & \checkmark & \textbf{68.4} & \textbf{53.5} & \textbf{37.2} & \textbf{484.8} & \textbf{0.618} \\
 \bottomrule
\end{tabular}
\end{center}
\end{table}

\paragraph{\bf{Analysis of DMA}} As mentioned in Section \ref{sec_decoder}, we explore the variants of $g_{k, i}$; softmax with temperature~\cite{hinton2015distilling} and (ST) Gumbel softmax~\cite{Jang2017}, namely, $g_{k, i}=0$ in softmax, and $g_{k, i}=-\log(-\log u), u \sim U(0,1)$ in (ST) Gumbel softmax in Eq. (\ref{eq_gumbel_zeta}).

Table \ref{Tab_Bengio} shows the results. We found that Gumbel softmax worked best. We assume this is because the stochastic sampling had a similar effect as ensembling and led to a slightly better performance. ST Gumbel softmax was found to be worse than the vanilla Gumbel softmax, which indicates that it is better to use a soft attention rather than a hard one. However, ST Gumbel softmax would be a good choice if it is desirable to completely switch off certain modalities, \eg for reducing sensor, computation, or memory cost.

\begin{figure}[tb]
\centering
\includegraphics[width=0.39\textwidth]{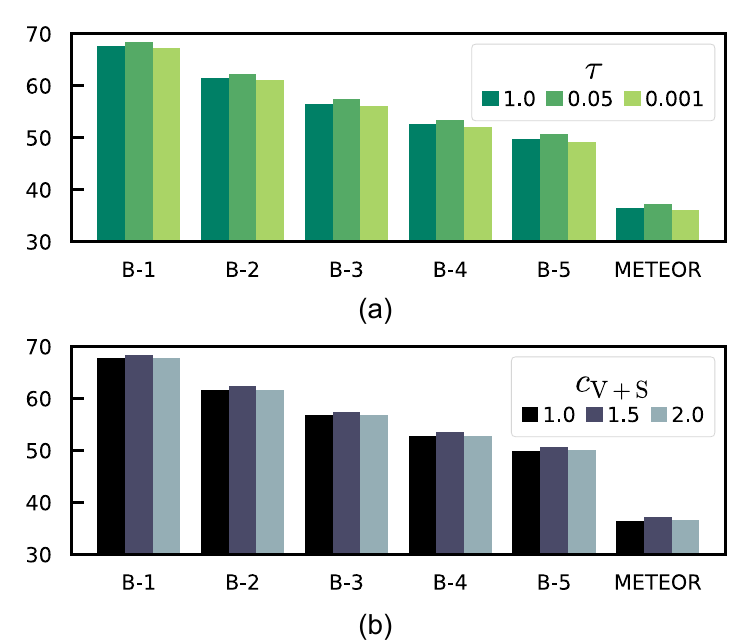}
\caption{Sensitivity analysis of hyper-parameters.}
\label{Fig_hyper_parameter_ablation}
\vspace{-4mm}
\end{figure}

Figure \ref{Fig_hyper_parameter_ablation} (a) shows the sensitivity of hyper-parameter $\tau$. We found that introducing a slightly hard attention by $\tau=0.05$ gave the best performance, whereas a too hard attention, namely $\tau=0.001$, degraded the performance. This result agrees with the aforementioned observation on Gumbel and ST Gumbel softmax.

Figure \ref{Fig_hyper_parameter_ablation} (b) shows the ablation for hyper-parameter $c_{k}$; a modality preference. We set $c_{\rm V}=c_{\rm S}=1.0$. The best performance was obtained on $c_{\rm V+S}=1.5$. The result suggests that it is better to slightly emphasize V+S over single modalities, which indicates the effectiveness of multi-modal information.

\begin{table}[tb]
\begin{center}
\caption{A comparison on weighted representation for DMA.}
\label{Tab_Bengio}
\begin{tabular}{@{}llllll@{}}
\toprule
Method & B-1 &B-4 & M & C & S \\ \midrule
Softmax w/ temperature & 67.9 & 52.8 & 36.5 & 474.8 & 0.612 \\
ST Gumbel softmax & 67.8 & 52.8 & 36.7 & 473.2 & 0.612 \\
Gumbel softmax & \textbf{68.4} & \textbf{53.5} & \textbf{37.2} & \textbf{484.8} & \textbf{0.618} \\ \bottomrule \end{tabular}
\end{center}
\end{table}

\begin{figure}[tb]
\centering
\includegraphics[width=0.41\textwidth]{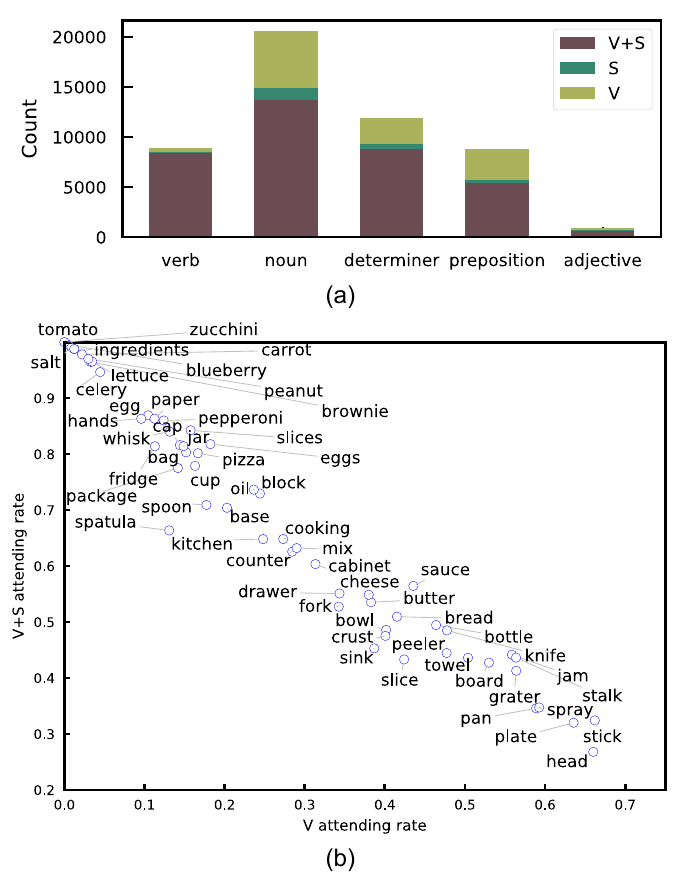}
\caption{Further analysis on DMA: (a) Number of word-types for V and V+S, (b) Scatter plot on attending rate for nouns.}
\label{Fig_DMA_analysis}
\vspace{-4mm}
\end{figure}

\paragraph{\bf{Detailed Analysis on Word-types}} Finally, we analyze the relationship between the generated word-types and modal attention. Figure \ref{Fig_DMA_analysis} (a) shows which modalities were attended to most, or ${\rm argmax}_{k} \zeta_{k, i}$, by word-types. We can see that $\mathbf{h}_{\rm V+S}$ was attended most in most cases. In particular, $\mathbf{h}_{\rm V+S}$ was utilized for generating verbs in 99.6\% of all the cases, suggesting the effectiveness of multi-modal information for describing motions. In addition, we confirmed that the visual representations $\mathbf{h}_{\rm V}$ were emphasized more often when generating nouns, prepositions, and determiners, with rates ranging from 26.2\% to 35.3\%. This indicates that sensor data are not always useful and it is better to focus only on visual representation if sensor data do not contain useful information or contain harmful information such as noise or signals of irrelevant motions. Figure \ref{Fig_DMA_analysis} (b) shows a scatter plot of the average attention rate of the modalities with respect to each noun words. The upper left region means that $\mathbf{h}_{\rm V+S}$ was almost always attended to most for generating the nouns in this region, and the lower right region means $\mathbf{h}_{\rm V}$ was used more often to generate the nouns in this region. We found that easily identifiable objects with only vision data, \eg {\it plate}, tended to be in the lower right region, while confusing objects, \eg ingredients, tended to be in the upper left. The result indicates that DMA worked reasonably from a qualitative perspective as well.

\section{Conclusion}
This paper proposed a novel task of egocentric multi-modal captioning, which incorporates wearable sensors to generate detailed activity description. To address this task, we constructed a dataset called MMAC Captions. We proposed the model for sensor-augmented video captioning, which has two key modules; AMMT and DMA. Experimental results demonstrated that AMMT enabled to effectively integrate multi-modal representations, DMA gave more precise captioning by appropriately attending to the preferable modalities, and the combination of these two resulted in consistently superior performance compared with the baselines. We believe this study will facilitate the further research and open up an opportunity for developing new applications such as automatic generation of detailed industrial-operation-manuals.

\begin{acks}
The CMU-MMAC data used in this paper was obtained from \url{kitchen.cs.cmu.edu} and the data collection was funded in part by the National Science Foundation under Grant No. EEEC-0540865.
\end{acks}

%\clearpage

\bibliographystyle{ACM-Reference-Format}

\balance

%\bibliography{ref_acm2021}


\end{document}